\documentclass{article}

\PassOptionsToPackage{numbers, compress}{natbib}

\usepackage[preprint]{neurips_2023}

\usepackage[utf8]{inputenc} 
\usepackage[T1]{fontenc}    
\usepackage{hyperref}       
\usepackage{url}            
\usepackage{booktabs}       
\usepackage{pifont}
\usepackage{amsmath}     
\usepackage{amsfonts}       
\usepackage{nicefrac}       
\usepackage{microtype}      
\usepackage{xcolor}         
\usepackage{algorithm}
\usepackage{algpseudocode}

\usepackage{graphicx}

\title{Augmenting Hessians with Inter-Layer Dependencies for Mixed-Precision Post-Training Quantization}

\author{%
    Clemens JS Schaefer\thanks{Work conducted while interning at Google LLC}$^{\;\;\dagger}$, Navid Lambert-Shirzad$^{\ddagger}$, Xiaofan Zhang$^{\ddagger}$, Chiachen Chou$^{\ddagger}$, \\
    \textbf{ Tom Jablin$^{\ddagger}$, Jian Li$^{\ddagger}$, Elfie Guo$^{\ddagger}$, Caitlin Stanton$^{\ddagger}$, 
      Siddharth Joshi$^{\dagger}$, and Yu Emma Wang$^{\ddagger}$}\\
    $^{\dagger}$ University of Notre Dame, Notre Dame, IN, USA \\
    $^{\ddagger}$ Google LLC, Mountain View, CA, USA \\
    \textit{cschaef6@nd.edu, yuemmawang@google.com}
}

\begin{document}

\maketitle

\begin{abstract}
Efficiently serving neural network models with low latency is becoming more challenging due to increasing model complexity and parameter count. Model quantization offers a solution which simultaneously reduces memory footprint and compute requirements. However, aggressive quantization may lead to an unacceptable loss in model accuracy owing to differences in sensitivity to numerical imperfection across different layers in the model. To address this challenge, we propose a mixed-precision post training quantization (PTQ) approach that assigns different numerical precisions to tensors in a network based on their specific needs, for a reduced memory footprint and improved latency while preserving model accuracy. Previous works rely on layer-wise Hessian information to determine numerical precision, but as we demonstrate, Hessian estimation is typically insufficient in determining an effective ordering of layer sensitivities. We address this by augmenting the estimated Hessian with additional information to capture inter-layer dependencies. We demonstrate that this consistently improves PTQ performance along the accuracy-latency Pareto frontier across multiple models. Our method combines second-order information and inter-layer dependencies to guide a bisection search, finding quantization configurations within a user-configurable model accuracy degradation range. We evaluate the effectiveness of our method on the ResNet50, MobileNetV2, and BERT models. Our experiments demonstrate latency reductions compared to a 16-bit baseline of $25.48\%$, $21.69\%$, and $33.28\%$ respectively, while maintaining model accuracy to within $99.99\%$ of the baseline model.
\end{abstract}

\section{Introduction}

Neural networks (NNs) underpin many machine learning (ML) systems, achieving state-of-the-art (SOTA) performance across a wide range of tasks, including computer vision~\cite{zhai2022scaling}, natural language processing~\cite{brown2020language}, and generative models for text~\cite{openai2023gpt4,thoppilan2022lamda} and images~\cite{ramesh2021zero}. However, these remarkable capabilities incur substantial compute and memory costs making these models challenging to deploy at scale while guaranteeing quality of service. These challenges are further exacerbated by the increasing proliferation of ML across tasks~\cite{jumper2021highly}. Overcoming these challenges requires resource efficient models that balance deployment costs against quality of service (QoS) metrics (e.g., latency and accuracy). Researchers have addressed this need using a variety of techniques including: hardware-efficient NN designs~\cite{tan2019efficientnet}, pruning~\cite{mishra2021accelerating}, and quantization~\cite{gholami2021survey}. Among these, quantization offers the simultaneous benefit of reducing the model footprint, enabling cheaper compute primitives, and reducing NN inference latency with the corresponding reduction in compute-energy.

By perturbing model parameters from their trained values, quantization can degrade model accuracy. Most often, this is mitigated by either incorporating quantization during the initial training or additional training of the NN with quantized parameters, collectively referred to in the literature as quantization aware training (QAT)~\cite{dong2020hawq, wang2019haq}. However, by updating model parameters and subsequent revalidation, and requiring training/finetuning data, QAT incurs significant overheads during model deployment. 
Post-training quantization (PTQ) aims to avoid this by determining quantization scales and rounding schemes either on a small calibration dataset or in a data-free manner, minimizing changes to model parameters. This trades off quantization complexity and model revalidation efforts against the accuracy of the quantized model~\cite{wu2020integer}.

Recognizing the benefits from model quantization, commercially available NN accelerators such as NVIDIA GPUs~\cite{nvidia2020a100,nvidia2023h100} or Google TPUs~\cite{jouppi2021ten,jouppi2023tpu} support quantized operations at various bit-widths, e.g. int4, int8, fp8, fp16, fp32 or fp64, to facilitate efficient NN inference. Maximally exploiting these hardware capabilities is challenging in practice because different NN layers and operations need to be configured to different bit-widths to best balance model accuracy against serving efficiency. Since the search space of all possible bit-width choices is exponential with the number of layers (or tensor slices for finer-grained approaches), this presents a significant challenge for rapidly deploying quantized NNs while guaranteeing QoS. QAT tackles that challenge by: (i) training bit-widths alongside other model parameters, given model size constraints~\cite{bhalgat2020lsq+, uhlich2019mixed}, (ii) using black-box reinforcement learning solutions to determine bit-widths~\cite{wang2019haq}, or (iii) using auxiliary metrics to reduce the search space~\cite{dong2020hawq}. The increased complexity associated with PTQ has typically resulted in research either: (i) ignoring mixed precision PTQ entirely and quantizing the model with a single bit-width~\cite{guo2022squant, frantar2022optimal}, (ii) used Pareto frontier methods based on Hessian sensitivity and model size~\cite{cai2020zeroq}, or (iii) used integer programming~\cite{hubara2021accurate, frantar2022optimal} to determine mixed precision configurations.

While the model fine-tuning involved in QAT may introduce computational overhead compared to the simpler PTQ techniques, PTQ often results in a prohibitive quality gap for the same level of model compression~\cite{jouppi2021ten}. This quality compromise prevents the broader adoption and deployment of PTQ-based quantization methods. Existing mixed precision PTQ approaches assume the numerical precision of different layers are independent, i.e., that the decision of quantizing one layer is made regardless of other layers. As we will show, this assumption results in sub-optimal quantization configurations. Prior efforts are unable to effectively select insensitive layers to quantize, with some quantizing more sensitive layers resulting in a poorer quality than what is acceptable in production.
The situation is further exacerbated when models are trained and deployed by different entities, where training accuracy is chosen without considering any subsequent quantization. In this case, any quantization-induced drop in accuracy might be intolerable.

\begin{figure*}[t]
\begin{center}
\centerline{\includegraphics[width=.93\textwidth, trim={0 .9cm 2.8cm .05cm},clip]{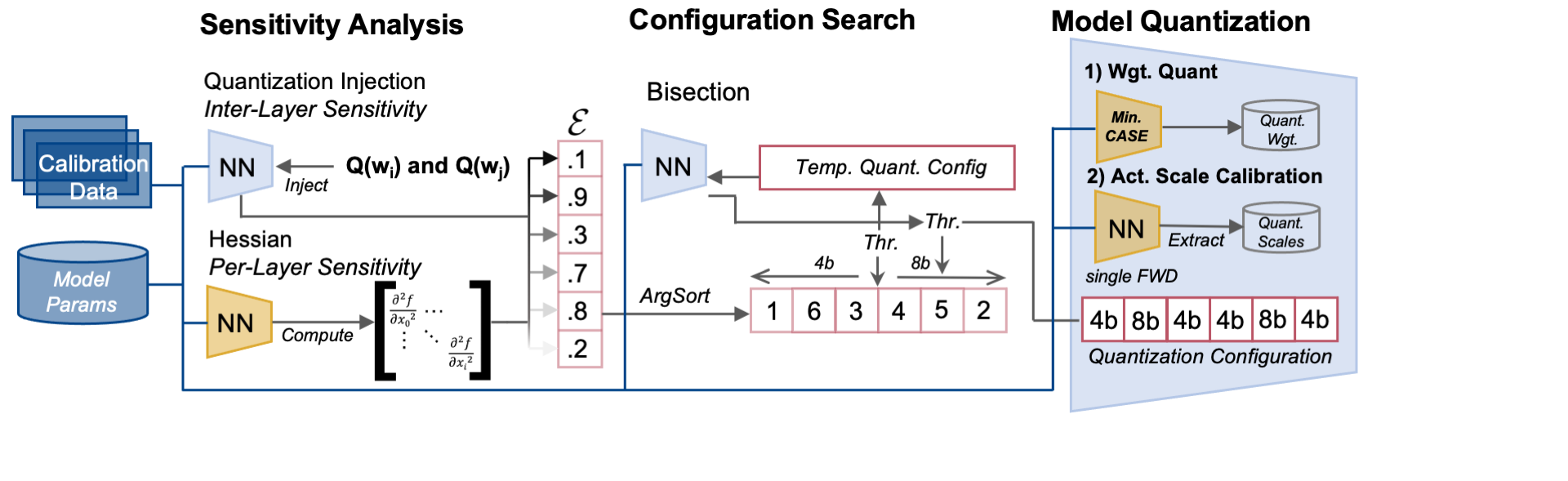}}
\vskip -0.1in
\caption{Summary of our approach. On the left we illustrate the fully-parallelizable computation of our sensitivity metric: the mean of the Hessian traces per layer and the loss after quantizing all pairs of layers. The Hessian for each matrix and inter-layer excess degradation are combined to guide a bisection search for the ideal bit allocation in the network given an accuracy target. On the right we illustrate our quantization method, for weights we follow SQuant~\cite{guo2022squant} and determine the ideal quantization rounding scheme data-free meanwhile for the activations we use a percentile calibration scheme to determine the quantization scales.}
\label{fig:meth}
\end{center}
\vskip -.4in
\end{figure*}

To tackle the challenges associated with mixed precision PTQ, this paper develops a unified method applicable to multiple model types (large scale, small scale, convolutional, and transformers) and data modalities (vision and text). Our approach enables deploying floating-point ML models to commercially available hardware with no manual intervention (see Figure~\ref{fig:meth}). As our primary contributions: (i) we demonstrate the ineffectiveness of the layer-wise sensitivity metric and introduce a novel metric that combines second-order information with inter-layer dependencies, (ii) we propose a guided bisection search to identify optimal quantization configurations while maintaining a production-level accuracy, and (iii) we evaluate our technique  experimentally on convolutional vision models and a transformer-based language model and show reductions of model footprints and inference latency given tight accuracy constraints. We demonstrate latency reductions of $25.48\%$ (ResNet50), $21.69\%$ (MobileNetV2), and $33.28\%$ (BERT) while maintaining model accuracy within 99.99\% of the baseline model on a calibration dataset.

\section{Related Work}

To find mixed precision quantization policies for QAT~\citet{wang2019haq} use reinforcement learning with feedback from a hardware accelerator , reporting a $1.4-1.95\times$ improvement to latency and $1.9\times$ improvement to power over a baseline eight-bit integer model with comparable accuracy. Gradient-based QAT to learn the precision on weights and activations have shown considerable success on models trained on the ImageNet task, with multiple competitive results on sub-5~MB models~\cite{schaefer2022edge,park2020profit}. When the numerical precision cannot directly be learned, alternative approaches typically employ a surrogate metric to determine layer importance or sensitivity and allocate precision accordingly. One example of such an approach was presented by \citet{yao2021hawq} propose to use the mean of the Hessian trace to determine layer sensitivity and develop a Pareto frontier of all model quantization configurations for use in QAT. They reduce the size of a ResNet50 to 7.99MB while achieving 75.76\% accuracy.

Due to the improvement to model compression observed during mixed-precision QAT, recent work has also studied the feasibility of applying mixed-precision quantization to PTQ. \citet{nahshan2021loss} investigate how quantization impacts the model loss landscape, observing flat separable structures for mild quantization and highly non-separable, steep curvature, for low bit-width quantization. Building on this they devise a three step method to improve PTQ: (i) determine the quantization step that minimizes a norm of the quantization error of the individual layers, (ii) use quadratic interpolation to approximate an optimum quantization scale, and (iii) jointly optimize the parameters of all layers acquired on the previous step by applying a gradient-free optimization method. In a similar way \citet{nagel2020up} theoretically analyze the impact of the rounding decision in during quantization and formulate it as a binary optimization problem (round up vs round down). Their proposed solution uses a layer-wise local loss, which can be optimized using a relaxation method for improved PTQ performance. \citet{yao2022zeroquant} demonstrate int4 and int8 PTQ on large Transformer-based models by using fine-grained quantization and layer-wise data-independent knowledge distillation. 

\citet{cai2020zeroq} introduce a mixed precision PTQ scheme that employs Hessian estimations, similar to previous QAT methods~~\cite{yao2021hawq}. To estimate the Hessian, the authors extract a distilled dataset from the unquantized model using batchnorm matching, which makes this inapplicable to transformer-based models. \citet{hubara2021accurate} quantize a model by updating its parameters to minimizes the error between the quantized layer output and full precision output, fine-tuning batchnorm parameters. They formulate the allocation of precision on a per-layer basis as an integer linear programming problem, with the cost being a function of the estimated model footprint and the accuracy. This method assumes strong inter-layer independence, and changes the model weights as well as batchnorm parameters, blurring the distinction between QAT and PTQ. This integer programming approach has also been adopted by other works such as~\cite{frantar2022optimal}. Alternative approaches to determining layer sensitivity have also been studied, such as signal-to-quantization noise~\cite{pandey2023practical} and Fisher Information~\cite{zandonati2022fit}. These metrics are used in a similar fashion, to construct an ordered list of layer sensitivity to facilitate a search for optimized mixed-precision quanization configurations.

To the best of our knowledge, \citet{zheng2022leveraging} are among few prior work examining inter-layer dependency for PTQ. They phrase the quantization process as a network-wise larger scale combinatorial optimization problem of discrete variables and enable efficient solution through various regularization techniques. However, they do not consider mixed precision and their accuracy degrades notably for the four bit configuration with eight bit top and bottom layers.

\section{Method}

Figure~\ref{fig:meth} provides and overview of our PTQ methodology. Initially, we determine per-layer sensitivity by approximating each layer's Hessian trace and inter-layer dependencies by assessing the impact of pairwise quantization across the model layers. We consolidate these into a single metric to establish an ordered sensitivity list. Next, we employ an bisection search to determine the sensitivity thresholds to facilitate a bit-width assignment to each layer that still meets the QoS requirements.  The right side of the figure illustrates the model quantization process, optimizing the rounding scheme for weights~\cite{guo2022squant}, and employing a percentile based calibration scheme for the activations~\cite{wu2020integer}.

\subsection{Quantization}

Fixed point quantization often termed integer quantization reduces the precision of numerical values in a model for a corresponding decrease in storage and compute requirements. This is typically achieved by applying clipping and rounding operations to the original floating-point values, often formulated as:
\begin{equation*}
Q(\mathbf{x}) = \text{round}(\text{clip}(\alpha \cdot \mathbf{x} ) \cdot 2^{b-1}) \cdot 2^{-(b-1)} \cdot \alpha^{-1}.
\end{equation*}
Here, $Q$ is the quantization function and $\mathbf{x}$ is the floating point value. The clipping function saturates values exceeding the thresholds to their corresponding extrema( minimum -1 and maximum 1), $b$ is the bit-width, and $\alpha$ is the quantization scale. 

To ensure compatibility with most commercially available hardware, we enforce that all operands (activations/weights) in a matrix multiply (matmul) have the same bit precision. For weights, we employ fine-grained quantization, where the rounding function and scale parameters are determined for each tensor dimension, e.g., per-channel, per-filter, or per-embedding. Building on previous work on PTQ, parameters remain unchanged and instead we adapt the rounding function~\cite{guo2022squant, li2021brecq, nagel2020up}. Our work minimizes the Constrained Absolute Sum of Error (CASE) by modifying the rounding direction for the values contributing the most to the CASE for that matmul~\cite{guo2022squant}. The scale ($\alpha$) for the weights are set based on the minimum and maximum observed along the tensor dimension. We determine a single scale for activations using single forward pass with a calibration set (a subset of elements from the training data). We employ a percentile-based method to determine the quantization scale for the activations~\cite{wu2020integer}, on a per-layer basis.

\subsection{Sensitivity Measures}
\begin{figure*}[t]
\centering
\includegraphics[width=0.85\textwidth]{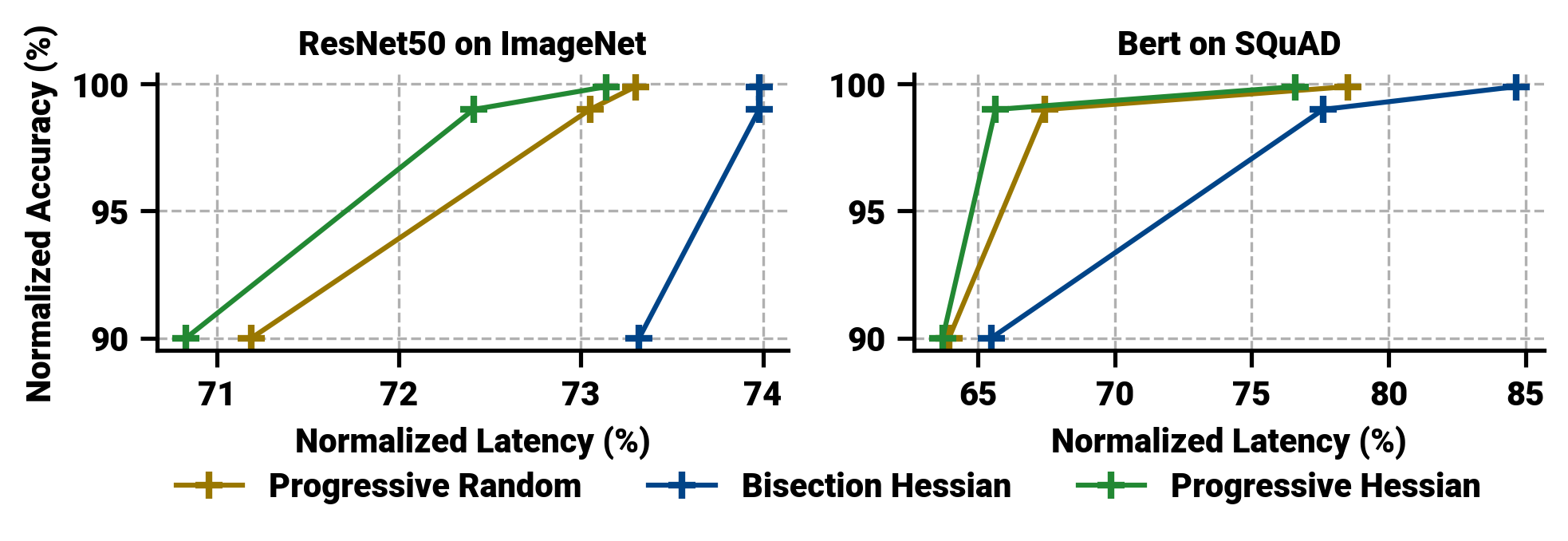} \vskip -.2in
\caption{The effectiveness of the Hessian trace as a sensitivity metric to guide quantization configuration search, comparing quantization outcomes for ResNet50 and BERT models: Hessian-guided bisection and progressive searches, and a random sensitivity-guided progressive search. The performance gap between Hessian-ordered bisection and progressive searches suggests that uncaptured interactions between multiple quantized layers are highly impactful. Though the progressive search can recover from misordered layer sensitivities, its runtime makes it impractical for production.}
\label{fig:greedy}
\end{figure*}
The space of possible configurations for a quantized model is exponential with the number tensors. Consider a ResNet50 with three different configuration options just for the parameters (e.g. bit-widths), this results in $3^{50}$ possible quantization configurations. Exhaustively evaluating these configurations is not practical for modern workloads. Consequently searching through this space efficiently is critical to deploying quantized models. The use of an informative sensitivity metric can reduce this vast space, making it practical to search for performant configurations.

One of the most commonly used sensitivity metric employs estimations of the Hessian, which pertains to the local curvature of a function. This choice is informed by theory that model accuracy is robust to perturbations in values that occupy flat regions of the loss function (low local curvature). However, for those values that occupy regions of high local curvature (sharp), small perturbations can have an exaggerated impact on model accuracy~\cite{dong2019hawq, rissanen1978modeling, hochreiter1997flat}. One way of estimating the local curvature, uses the Hessian of the loss function, which comprises second-order partial derivatives of the loss.

Rather than directly evaluating the Hessian, which is computationally prohibitive, we approximate the trace using Hutchinson's algorithm as seen in related work~\cite{dong2020hawq,lee2021network}. We define a Hessian-based metric for the $i$ layer of a network as:
\begin{equation}
\mathcal{E}_i^{\text{Hessian}} = \mathbf{E} \left[\mathrm{tr} \left(\frac{L(\mathbf{x}, \mathbb{W})}{\partial \mathbf{w}_i^2}\right)\right].\nonumber
\end{equation}
Where $\mathrm{tr}$ is the trace operator, $L$ the model's loss function, $\mathbb{W}$ the set of all considered tensors (e.g., weights/activations) and $\mathbf{x}$ the calibration data. Higher $\mathcal{E}_{\text{Hessian}}$ values signify increased local curvature of the loss function, implying greater model sensitivity to parameter changes. Sorting by $\mathcal{E}_{\text{Hessian}}$ gives an ordering of the ease of layer quantization.

Given an accurately ordered layer sensitivity list, a bisection search-like method can efficiently determine layer quantization configurations. However, as shown in Figure~\ref{fig:greedy}, a bisection search yields subpar results with layers ordered by the Hessian compared to a sequential (progressive) search algorithm. The progressive search, sequentially evaluates the suitability of assigning each layer a bit-width using cumulative model degradation as the assignment criterion (see Supplementary Materials~\ref{alg:greedy} for pseudo-code). We test for two orders in which layers are evaluated, first prioritized by the sensitivity metric $\mathcal{E}_{\text{Hessian}}$ or randomly ordered. We attribute the performance discrepancy between progressive and bisection searches to incorrect ordering of high sensitivity layers. This misclassification changes ordering and is recoverable by the progressive search but catastrophic for the bisection method. As Figure~\ref{fig:greedy} demonstrates, the performance of progressive search is competitive with the Hessian-guided search even with a randomly ordered sensitivity list, significantly outdoing the Hessian-guided bisection search. However, with a correctly ordered sensitivity list, both search methods should yield identical configurations.

\begin{figure}[t]
\centering
\includegraphics[width=0.9\textwidth, trim={0cm 0cm 0cm .2cm},clip]{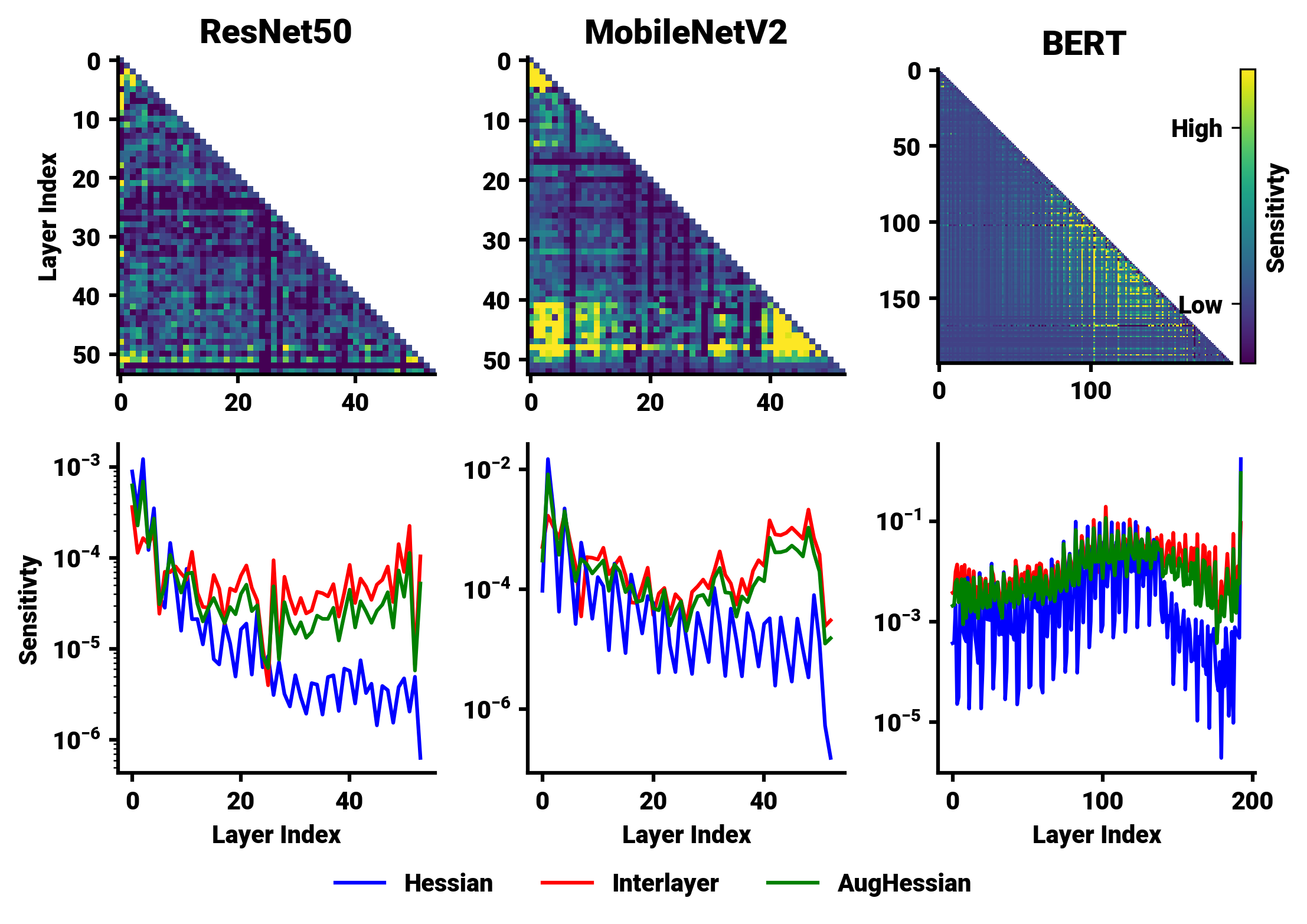}
\vskip -.2in
\caption{Sensitivity Metrics for ResNet50, MobileNetV2 and BERT. Top row shows the excess degradation of layer quantization combinations at eight bit. The convolutional networks exhibit high excess degradation at near the early and late layers, whereas the transformer model shows higher excess degradation towards the middle of the network. The bottom row plots the difference between the layer sensitivity obtained from the Hessian, the excess degradation, and augmented Hessian.}\vspace{-.25cm}
\label{fig:sens}
\end{figure}

Since assuming layer-wise independence does not produce an accurate ordering of the final per-layer sensitivity, we augment our sensitivity metric by estimating pairwise-layer sensitivities. Second order methods become cost prohibitive for this and Hutchinson's algorithm only captures the impact of the diagonal elements of the Hessian, making it unsuitable for our needs. Instead, we estimate multi-layer dependencies by directly quantizing the layers in a pairwise fashion:
\begin{align*}
    \mathcal{E}_i^{\text{InterLayer}} &= \sum^l_j L(\mathbf{x}, \mathbb{W}^{i,j}) - \max( L(\mathbf{x}, \mathbb{W}^i), L(\mathbf{x}, \mathbb{W}^j)),\\
    \mathbb{W}^{i,j} &= \left\{\mathbb{W} \setminus{\left\{\mathbf{w_i}, \mathbf{w_j}\right\}}, Q(\mathbf{w_i}),  Q(\mathbf{w_j}) \right\}.
\end{align*}
Here, we sum the excess degradation incurred given the interaction between two layers. We define excess degradation to mean the difference in the loss ($L$) between the jointly quantized loss and the single layer quantized loss per-layer. We clip the minimum $\mathcal{E}_i^{\text{InterLayer}}$ to 0, disregarding any negative values. We then normalize and scale  $\mathcal{E}_i^{\text{InterLayer}}$ to combine it with the $\mathcal{E}_i^{\text{Hessian}}$ as:
\begin{equation}
\mathcal{E}_i^{\text{AugHessian}} = \mathcal{E}_i^{\text{Hessian}} + \beta \mathcal{E}_i^{\text{InterLayer}}, \;\;\;\;\;\;\;\;\;
\beta =\frac{\mathbf{E}[\mathcal{E}_i^{\text{Hessian}}]}{\mathbf{E}[\mathcal{E}_i^{\text{InterLayer}}]}. \nonumber
\end{equation}

Figure~\ref{fig:sens} visualizes these metrics as well as their combination. Figure~\ref{fig:sens} (top) shows the excess degradation from quantizing NN layers, pairwise, for three models. For both the vision models, the early and late layers show a higher sensitivity, while the transformer based BERT exhibits higher sensitivity towards the center. As seen, the Hessian-based sensitivity measure does not result in the same ordering, where e.g., the impact of quantizing the last layer is underestimated. 

\subsection{Search for Quantization Configurations}

We use a bisection method to determine the quantization configuration with $\mathcal{O}(b\log{N})$ model evaluations. Here, $N$ is the total number of layers and $b$ the number of available quantization bit-widths. We implement this search on a sensitivity list sorted based on the augmented sensitivity measure ($\mathcal{E}^{\text{AugHessian}}$), to determine the threshold sensitivity value corresponding to different quantization levels. We evaluate the quantized configuration using the same calibration set used to determine scale parameters. The bisection search iteratively updates the threshold value, and thereby the quantization configuration, by expanding or decreasing the number of quantized layers depending on if the accuracy target is achieved. We progressively determine the sensitivity threshold for each available precision setting, starting form the highest (e.g., 8-bits) to the lowest (e.g., 4-bits). Pseudocode for the bisection algorithm is provided in Supplementary Materials Algorithm~\ref{alg:bisect}. 

\section{Experiments}

\begin{figure}[t]
\begin{center}
\centerline{\includegraphics[width=.95\columnwidth, trim={0 0 0 .2cm},clip ]{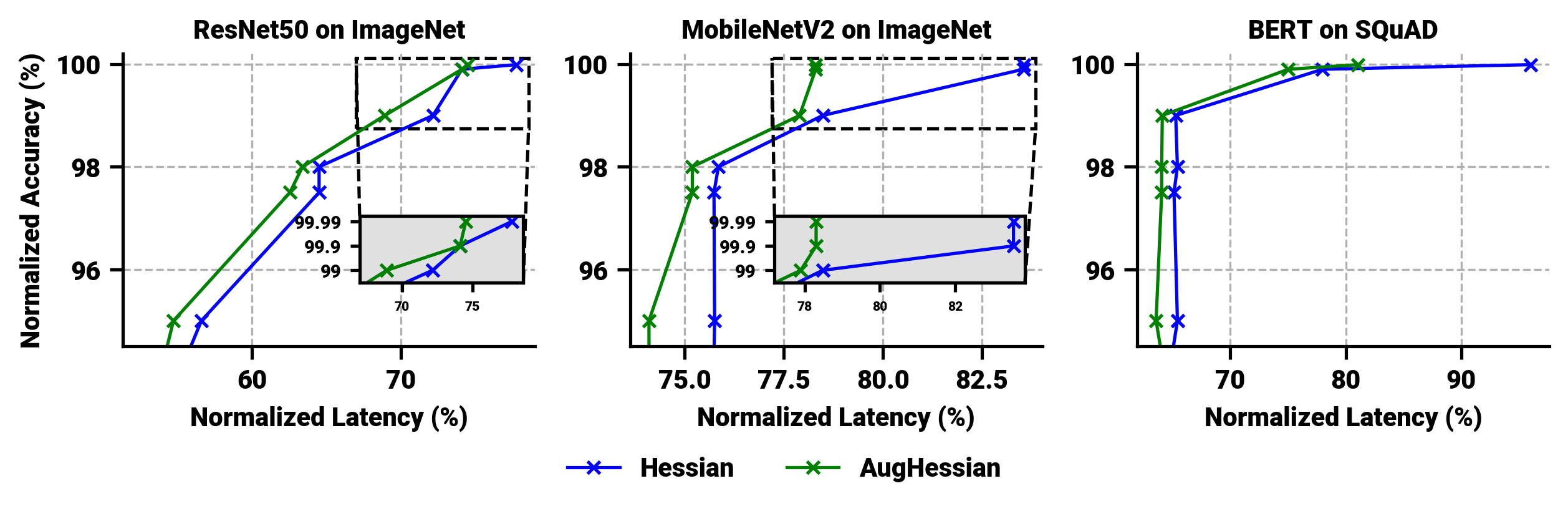}}\vspace{-.5cm}
\caption{Evaluation of our method on ResNet50, MobileNetV2, and BERT. Highlighting the performance gain of our proposed Augmented Hessian sensitivity metric over the Hessian, especially note the big latency gains at high accuracy levels of the MobileNetV2 and BERT model, suggestion our metric contains high-quality knowledge for quantization.}\vspace{-.25cm}
\label{fig:over}
\end{center}
\end{figure}

We evaluate our proposed method on the ImageNet~\cite{ILSVRC15} and SQuaAD~\cite{2016arXiv160605250R} datasets, using ResNet50~\cite{he2016deep}, MobileNetV2~\cite{sandler2018mobilenetv2}, and BERT~\cite{devlin2018bert}. ResNet50 (ImageNet) and BERT (SQuAD) are commonly accepted dataset-model combinations from the MLPerf inference suite~\cite{reddi2020mlperf}\footnote{\url{https://mlcommons.org/}}. We show results on MobileNetV2 to demonstrate the versatility of our method and performance on small edge models. For calibration, determining the sensitivity, and guiding the search we randomly sample 4096 examples from the original training data which we use for all steps. For activation calibration we set quantization scales based on the 99.999 percentile value observed during a forward pass of the calibration set through a model with quantized weights. We also improvements to the quantization performance for MobileNetV2 on adding a layer size penalty.

We estimate latency by benchmarking key kernels like \texttt{gemm} and \texttt{conv2d} at various numerical precisions on A100 GPUs, using an inference batch-size of one. Directly capturing the interplay between memory hierarchy, bus-speeds, compute-utilization, and compiler optimizations. We identified the top-performing kernels for specific tensor shapes and precisions using the CUTLASS~\cite{Kerr_CUTLASS_2022} profiler and optimizer. This data was then used to estimate deployment latencies for different multi-precision models. Our results (Tables~\ref{tab:resnet50},~\ref{tab:mbv2}, and~\ref{tab:bert}) show linear model size reduction with bit quantity and reflect the complex deployment interactions arising from latency reductions.

\setlength{\tabcolsep}{4pt}
\begin{table}[t]
\caption{Comparison of our results to other work for \textbf{ResNet50}. Our method (augmented Hessian with 99.9\%) achieves highest accuracy while enabling 25.88\% latency reduction through quantization over 16 bit model. See text for explanation of annotation marks in the table.} \vspace{-.25cm}
\label{tab:resnet50}
\begin{center}
\begin{small}
\begin{tabular}{@{}lccccccccccccc@{}}
\toprule
\textbf{ResNet50} & & &  \multicolumn{2}{c}{Accuracy} & & \multicolumn{2}{c}{Size} & & \multicolumn{2}{c}{Latency} & & \multicolumn{2}{c}{Precision} \\
& No FT & & Absolute & Relative$^*$ & &  MB & Relative & & ms & Relative & & W & A\\
\cmidrule{2-2} \cmidrule{4-5} \cmidrule{7-8} \cmidrule{10-11} \cmidrule{13-14} 
Baseline (ours) & & & 77.60 & 100.00\% & & 51.00 & 100.00\% & & 5.20  & 100.00\% & & 16 & 16  \\
\midrule
MrBiQ \cite{jeon2022mr} & \checkmark & & 75.17 & 97.62\%  & & 13.78 & 27.02\%  & & 2.70  & 51.87\%   & & 4 & 4$^\dagger$ \\
ZeroQ \cite{cai2020zeroq} & \checkmark & & 76.08 & 97.89\%  & & 12.73 & 24.97\%  & &  3.82  & 73.46\%  & & MP & 8 \\
QDrop \cite{wei2022qdrop} & \checkmark & & 75.45 & 97.99\%  & & 13.78 & 27.02\%  & & 2.70  & 51.87\%  & & 4 & 4$^\dagger$ \\
LAPQ \cite{nahshan2021loss} & \checkmark & & 74.80 & 98.29\%  & & 25.50 & 50.00\%  & & 3.82  & 73.46\%  &  & 8 &4  \\
AdaQuant$^\ddagger$ \cite{hubara2021accurate} & x$^\star$ & & 75.90 &  98.32\%  & & 24.32 &  47.69\% & & 3.79  & 72.88\%  & & MP & MP \\
AdaRound \cite{nagel2020up} & \checkmark & & 75.01 & 98.61\%  & & 25.50 & 50.00\%  & & 3.82  & 73.46\%  & & 4 & 8 \\
HAWQV3 \cite{yao2021hawq} & x & & 76.73 & 98.73\%  & & 18.70 & 36.67\%  & & 3.28  & 63.16\%  & & MP & MP \\
BSQ \cite{yang2021bsq} & x & & 75.29 & 98.90\%  & & - & - &  & 2.70  & 51.87\%  & & MP & 4$^\dagger$ \\
OBQ \cite{frantar2022optimal}  & x$^\star$ & & 75.72 & 99.17\%  & & 12.75 & 25.00\%  & & 2.68  &  51.54\%  & & 4 & 4  \\
PTQMP \cite{pandey2023practical} & \checkmark & & 75.95  & 99.76\%  & & 25.50 & 50.00\%  & & 3.82  & 73.46\%  & & MP & MP  \\
SQuant \cite{guo2022squant} & \checkmark & & 77.66  & 99.91\%  & & 25.50 & 50.00\%  & & 3.82  & 73.46\%  & & 8 & 8  \\
\midrule
Hessian 99\% &\checkmark & & 77.13 & 99.39\%  & & 21.54 & 42.24\%  & & 3.75  & 72.17\%  & & MP & MP  \\
AugHessian 99\% & \checkmark & & 77.19 & 99.46\%  & & 23.68 & 46.43\%  & & 3.58  & 68.93\%  & &  MP & MP  \\
Hessian 99.9\%  & \checkmark & & 77.26 & 99.55\% & & 25.52 & 50.03\%  & & 3.85  & 74.12\%  &  & MP & MP  \\
AugHessian 99.9\% & \checkmark & & 77.57 & 99.95\%  & & 25.52 & 50.03\% &  & 3.85  & 74.12\%  & & MP & MP  \\
\bottomrule
\end{tabular}\vspace{-.5cm}
\end{small}
\end{center}
\end{table}

\setlength{\tabcolsep}{4pt}
\begin{table}[t]
\caption{\textbf{MobileNetV2} results and comparison to other work. Previous work do not reach accuracy within 1\% of the baseline model. Our method offers multiple configuration within 1\% degradation which offer up to 22.10\% latency reductions compared to unquantized models. See text for explanation of annotation marks in the table.
}\vspace{-.25cm}
\label{tab:mbv2}
\begin{center}
\begin{small}
\begin{tabular}{@{}lccccccccccccc@{}}
\toprule
\textbf{MobileNetV2} & & &  \multicolumn{2}{c}{Accuracy} & & \multicolumn{2}{c}{Size} & & \multicolumn{2}{c}{Latency} & & \multicolumn{2}{c}{Precision} \\
& No FT & & Absolute & Relative$^*$ & &  MB & Relative & & ms & Relative & & W & A\\
\cmidrule{2-2} \cmidrule{4-5} \cmidrule{7-8} \cmidrule{10-11} \cmidrule{13-14} 
Baseline (ours) & & & 71.52 & 100.00\% & & 6.94 & 100.00\% & & 3.97  & 100.00\% & & 16 & 16  \\
\midrule
LAPQ \cite{nahshan2021loss}  & \checkmark & &  65.10 & 90.67\%  & & 1.73  & 25.00\% & & 3.97  & 100.00\% & & 4  & 32 \\
BRECQ \cite{li2021brecq} & \checkmark & &  66.57 & 91.83\%  & & 2.38  & 34.23\%  & & 2.55  & 64.18\%  & & 4 & 4$^\ddagger$ \\
QDrop \cite{wei2022qdrop} & \checkmark & &  68.84 & 94.96\%  & & 2.38  & 34.23\%  & & 2.55  & 64.18\%  & & 4 & 4$^\dagger$ \\
ZeroQ \cite{cai2020zeroq} & \checkmark & &  69.44 & 95.08\%  & & 0.87  & 12.49\%  & & 3.02  & 76.05\%  & & MP  & 8 \\
MrBiQ \cite{jeon2022mr} & \checkmark & &  68.97 & 95.14\%  & & 2.38  & 34.23\%  & & 2.55  & 64.18\%  & & 4 & 4$^\dagger$ \\
NWQ \cite{zheng2022leveraging} & \checkmark & &  69.60 & 96.01\%  & & 2.38  & 34.23\%  & & 2.55  & 64.18\%  & & 4 & 4$^\dagger$ \\
AdaQuant$^\ddagger$ \cite{hubara2021accurate}   & x$^\star$ & &  70.22 & 96.14\%  & & 5.09  & 73.29\%  & & 3.18  & 80.22\%  & & MP  & MP \\
AdaRound \cite{nagel2020up}  & \checkmark & &  69.25 & 96.56\%  & & 1.73  & 25.00\%  & & 3.02  & 76.05\%  & & 4 & 8 \\
DFQ \cite{nagel2019data} & \checkmark & &  71.20 & 97.49\%  & & 3.47  & 50.00\%  & & 3.02  & 76.05\%  & & 8 & 8 \\
PTQMP \cite{pandey2023practical}  & \checkmark & &  70.68 & 98.34\%  & & 3.47  & 50.00\%  & & 3.02  & 76.05\%  & & MP & MP \\
PTQ-MP \cite{liu2021post}   & \checkmark & &  70.70 & 98.50\%  & & 3.47  & 50.00\%  & & 3.02  & 76.05\%  & & 8 & 8 \\
\midrule
Hessian 99\% &\checkmark & & 71.01 & 99.28\%  & & 3.47 & 50.02\%  & & 3.12  & 78.48\%  & & MP & MP  \\
AugHessian 99\% & \checkmark & & 71.25 & 99.61\%  & & 4.75 & 68.44\%  & & 3.09  & 77.90\%  & &  MP & MP  \\
Hessian 99.9\%  & \checkmark & & 71.20 & 99.56\% & & 3.49 & 50.26\%  & & 3.32  & 83.55\%  &  & MP & MP  \\
AugHessian 99.9\% & \checkmark & & 71.34 & 99.75\%  & & 4.75 & 68.47\% &  & 3.11  & 78.31\%  & & MP & MP  \\
\bottomrule
\end{tabular}
\end{small}
\end{center}
\end{table}

\setlength{\tabcolsep}{3pt}
\begin{table}[t]
\caption{Our results on \textbf{Bert} and comparison to other work. Comparable other works do not leverage mixed precision, by doing so we deliver best absolute and relative accuracy while also beating latency of other works operating at eight bits. See text for explanation of annotation marks in the table.}\vspace{-.25cm}
\label{tab:bert}
\begin{center}
\begin{small}
\begin{tabular}{@{}lcccccccccccccc@{}}
\toprule
\textbf{BERT} & & &  \multicolumn{2}{c}{Accuracy} & & \multicolumn{2}{c}{Size} & & \multicolumn{2}{c}{Latency} & & \multicolumn{3}{c}{Precision} \\
& No FT & & Absolute & Relative$^*$ & &  MB & Relative & & ms & Relative & & W & A & E\\
\cmidrule{2-2} \cmidrule{4-5} \cmidrule{7-8} \cmidrule{10-11} \cmidrule{13-15} 
Baseline (ours) &  & &  90.25 & 100.00\% & & 603.98  & 100.00\% & & 4.28  & 100.00\% & & 8 & 8 & 8 \\
\midrule
QDrop \cite{wei2022qdrop} & \checkmark & &  77.26 & 87.38\%  & & 151.00  & 25.00\%  & & 2.31  & 53.93\%  & & 4 & 8 & 4 \\
QBert \cite{shen2020q} & x & &  86.95 & 98.04\%  & & 151.00  & 25.00\%  & & 2.79  & 65.19\%  & & 8 & 8 & 8 \\
MREM-P \cite{bai2022towards}  & \checkmark & &  87.30 & 98.42\%  & & 151.00  & 25.00\%  & & 2.79  & 65.19\%  & & 8 & 8 & 8 \\
BRECQ \cite{li2021brecq} & \checkmark & &  87.41 & 99.08\%  & & 301.99  & 50.00\%  & & 2.79  & 65.19\%  & & 8 & 8 & 8 \\
Q8Bert \cite{zafrir2019q8bert} & x & &  87.74 & 99.19\%  & & 301.99  & 50.00\%  & & 2.79  & 65.19\%  & & 8 & 8 & 8 \\
MrBiQ \cite{jeon2022mr}  & \checkmark & &  87.69 & 99.40\%  & & 113.25  & 18.75\%  & & 2.79  & 65.19\%  & & 8 & 8 & 8 \\
AdaQuant$^\ddagger$ \cite{hubara2021accurate}   & x$^\star$ & &  88.70 & 99.88\%  & & 301.99  & 50.00\%  & & 2.79  & 65.19\%  & & 4 & 8 & 8 \\
\midrule
Hessian 99\% & \checkmark & &  88.94 & 98.54\%  & & 295.18  & 48.87\%  & & 2.79  & 65.29\%  & & MP  & MP  & MP \\
AugHessian 99\%  & \checkmark & &  89.56 & 99.23\%  & & 285.74  & 47.31\%  & & 2.74  & 64.13\%  & & MP  & MP  & MP \\
Hessian 99.9\% & \checkmark & &  90.24 & 99.98\%  & & 435.16  & 72.05\%  & & 3.34  & 78.00\%  & & MP  & MP  & MP \\
AugHessian 99.9\%  & \checkmark & &  90.22 & 99.96\%  & & 389.55  & 64.50\%  & & 3.21  & 75.03\%  & & MP  & MP  & MP \\
Hessian 99.99\%  & \checkmark & &  90.29 & 100.04\% & & 579.87  & 96.01\%  & & 4.11  & 96.00\%  & & MP  & MP  & MP \\
AugHessian 99.99\% & \checkmark & &  90.24 & 99.98\%  & & 434.64  & 71.96\%  & & 3.47  & 81.06\%  & & MP  & MP  & MP \\
\bottomrule
\end{tabular}\vspace{-.5cm}
\end{small}
\end{center}
\end{table}

We present our experimental results and contextualize them with respect to other work in Tables~\ref{tab:resnet50}, \ref{tab:mbv2}, and \ref{tab:bert}, summarizing absolute accuracy, model size, latency, as well as relative measures. The tables provide results for two search settings: a 99\% and 99.9\% accuracy target (and 99.99\% for BERT). Our method delivers competitive model latency and model compression while exceeding the accuracy delivered by other techniques that result in similarly compressed models. Some entries in the table have additional annotations, for fair comparison across techniques. We distinguish between QAT and PTQ by indicating the presence or absence of finetuning (\texttt{No FT}). We always use the reported baseline indicating relative drop in accuracy using $^*$. We use $^\dagger$ to indicate that models quantized first and last layers to 8 bit, $^\ddagger$ to indicate a manual rerun of \textit{light pipeline}, and $^\star$ to indicate that the method implements batch-norm finetuning. Across all search targets, the augmented Hessian sensitivity metric consistently improves upon model latency compared to the pure Hessian metric. For ResNet50, targeting a 99.9\% accuracy, our quantized model attained the highest accuracy with 99.9\% on the calibration set and 99.95\% on the complete ImageNet validation set, while delivering a 25.88\% reduction in latency. Similar outcomes were observed for MobileNetV2, where no other model exceeded the 99\% accuracy threshold compared to the unquantized baseline. Our method delivered up to 99.75\% accuracy while reducing latency by 21.69\%. When quantizing BERT models, we observed a 16.93\% latency difference between the accuracy targets of 99\% and 99.99\%, achieving the highest accuracy among quantized models while still improving model serving latency.

\begin{figure*}[ht]
\begin{center}
\centerline{\includegraphics[width=.9\textwidth]{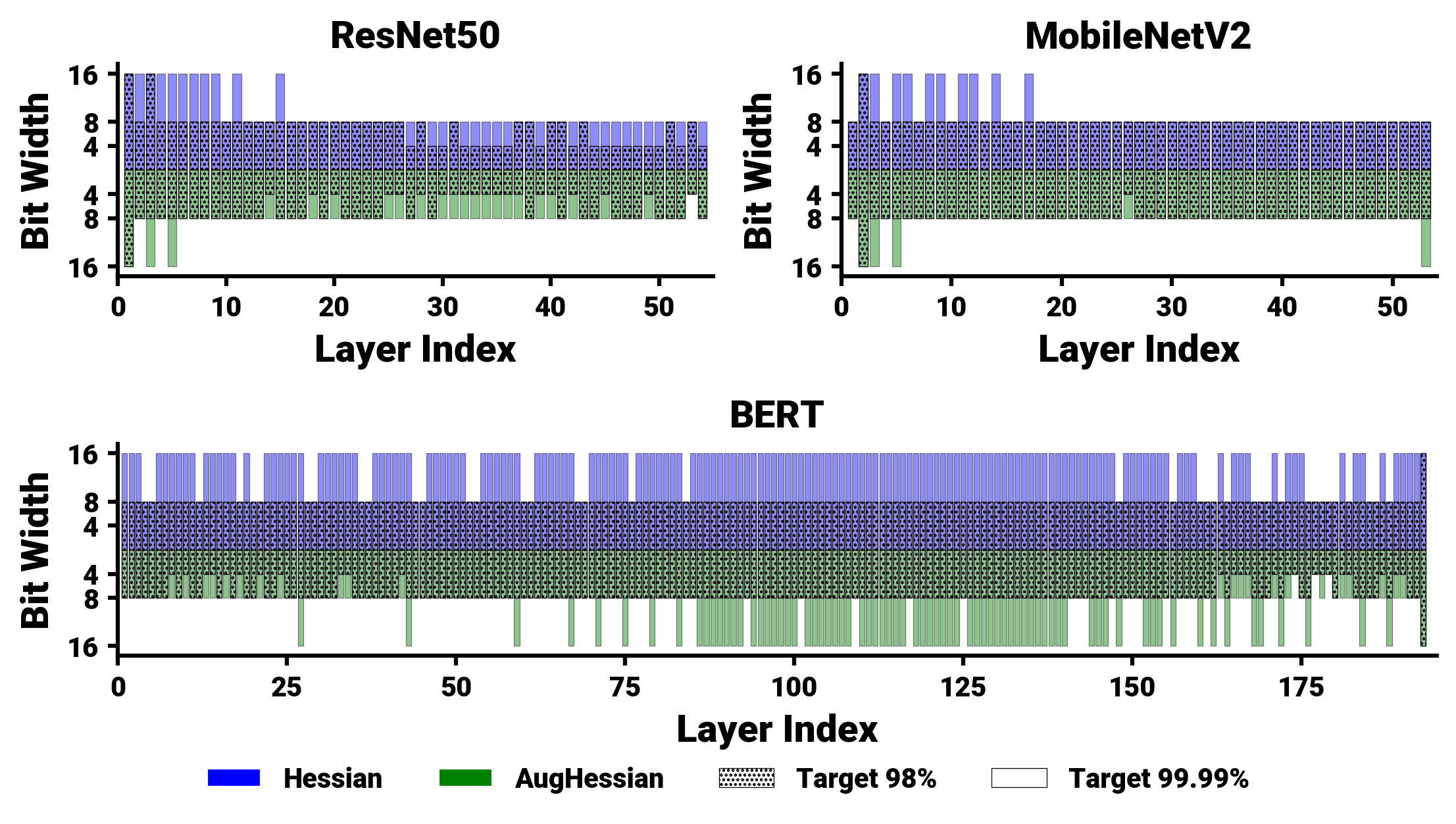}}
\vskip -0.2in
\caption{Per-layer bit-width configurations for ResNet50, MobileNetV2, and BERT. The blue bars are bit-widths obtained using a Hessian sensitivty metric and green bars with augmented Hessians, with stippling distinguishing the 98\% accuracy target from 99.99\%. The convolutional networks can be quantized to 8-bits, with some sensitive, 16-bit, layers distinguishing between the two sensitivity metrics. Relaxing the accuracy target to 98\% results in more four bit layers. For BERT, the augmented Hessian metric can more aggressive quantize the early and latter layers in the network, with most layers quantized to 8-bits for the 98\% target accuracy.}\vspace{-.5cm}
\label{fig:comp}
\end{center}
\end{figure*}

Figure~\ref{fig:over} shows the performance difference arising from the use of the two sensitivity metrics. For all configurations, the model derived from the augmented Hessian occupies a superior position on the accuracy-latency frontier. Ablation studies show the impact of using only inter-layer dependency to guide search (Supplementary Materials section~\ref{sec:abl}). While relaxed target accuracies the two metrics result in similar model serving latency, with more stringent targets the augmented search metric improves upon the Hessian by 5-15\% across all models. We provide more insight on the difference between the two sensitivity metrics in Figure~\ref{fig:sens1} in Supplementary Materials. Figure~\ref{fig:comp} shows a detailed bit allocation break down for all three models. The major difference between using only the Hessian for search guidance vs. the augmented Hessian is that more layerer are quantized to eight bits, especially visible for early layers. Additionally we show the difference between a 98\% and 99.99\% accuracy target, which manifests itself as more layers quantized to four bits for the augmented Hessian sensitivity. Table~\ref{tab:eval_num} in the Supplementary Materials shows how many evaluations our bisection search took for the 99\% and 99.9\% accuracy targets in Figure~\ref{fig:over}, the values are aligned with the theoretical expectations of $\mathcal{O}(N)$ where $N$ is the number of layers. With an average of only six evaluations our bisection search is significantly faster than sequential search.

\paragraph{Limitations}
We have not evaluated the impact of mixed-precision kernels e.g., 4W8A, but we do not foresee complications arising from this for our approach. Our latency estimates are currently pessimistic since they do not capture the impact of kernel/operator fusion. Computing the excess degradation requires $\frac{l \cdot (l - 1)}{ 2} + l$ ($l$ number of layers) model evaluations. While these can be parallelized and batched, these evaluations might still be costly for larger models. Additionally, estimating the trace of the Hessian remains a computationally intensive task.  Although this has not limited us for the models we evaluated, but might impact applicability for significantly larger models. Because we use a calibration dataset at multiple steps,  quantization performance will strongly depend on the alignment between the calibration data and the evaluation/real world data. Additional research is needed to implement this in a completely data-free fashion~\cite{zhong2022intraq}.

\section{Conclusions}
We introduce a practical mixed precision PTQ pipeline for efficiently quantizing floating point NN models while maintaining a target accuracy on a calibration dataset. Our technique calibrates the quantizer scales and adapts the weight rounding scheme but does not adapt any of the original model parameters (including batch-norm parameters). We demonstrate the limitations of assuming layer independence in estimating layer sensitivity and address this using a new sensitivity metric that also captures the pairwise interaction between multiple quantized layers. This improved metric enables us to use a bisection-search to determine quantization configurations that outperform the unaugmented sensitivity metric. Our method is demonstrated across small (MobileNetV2), medium (ResNet50), and large-scale (BERT) models, applicable to both vision and text data modalities. It achieves latency improvements ranging from 25-33\% with minimal impact on accuracy. On an anverage, we require six model evaluations to find these quantization configurations across the tested models.

\paragraph{Broader Impacts}
Our mixed-precision PTQ can deploy models with low latency creating positive impacts such as improved accessibility or reduced energy consumption (contributing to sustainability and cost-effectiveness). However we also consider risks of negative impacts such as limited interpretability mostly due to a lack of research on the interpretablilty of quantized models (reducing transparency and trust) and potential bias and fairness issues which are actively researched~\cite{hooker2020characterising}.

\bibliographystyle{plainnat}
\bibliography{references}

\newpage
\appendix
\section{Supplementary Materials}

\subsection{Error Bars}

\begin{figure}[H]
\centering
\includegraphics[width=1.0\textwidth]{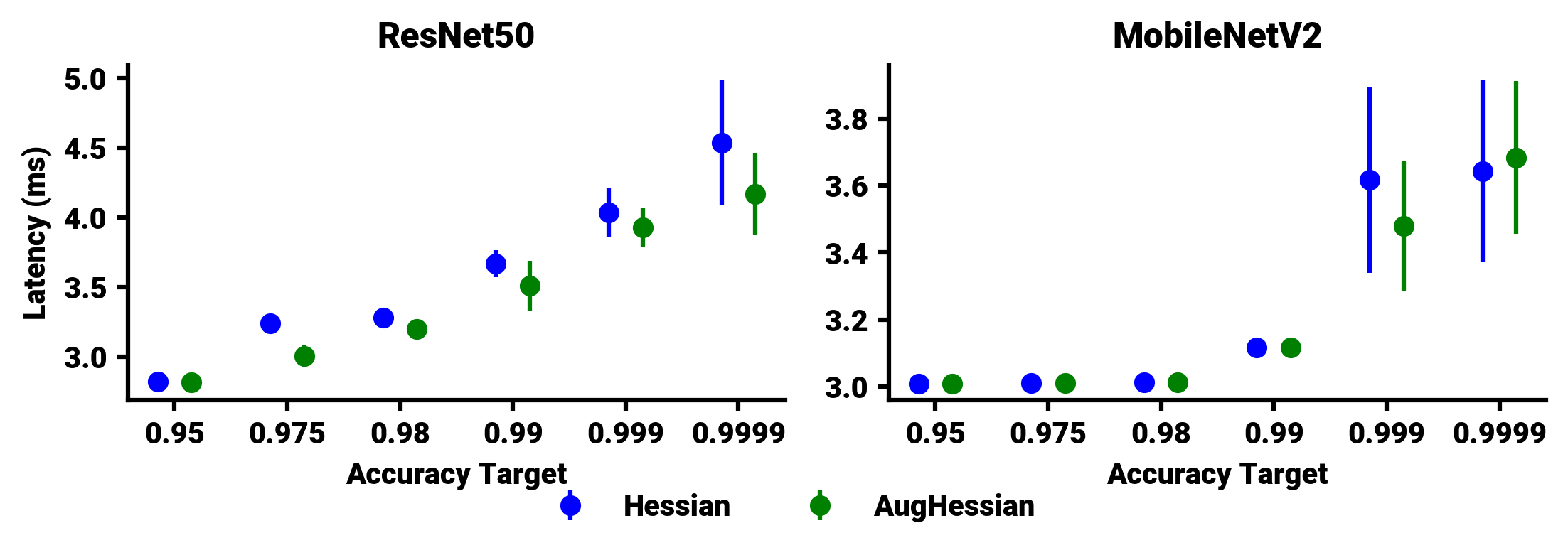}
\caption{Error bars over five trials for ResNet50 and MobileNetV2}
\label{fig:errb}
\end{figure}

In Figure~\ref{fig:errb} we show error bars over five trials where we changed the calibration and evaluation data of our bisection search to analyze the variance of our results. For ResNet50 we show that augmented Hessians are consistently better than pure Hessian information where the standard deviation is increasing which higher accuracy targets. For the MobileNetV2 the story roughly holds true however the latency mean using Hessian information for the highest accuracy target is lower. We contributed that the higher variance where by small uninformed random perturbation result in better models.

\subsection{Greedy Search} \label{sec:abl}

We also use the Hessian and our augmented Hessian in combination with a greedy search method (outlined in ~\ref{alg:greedy}) to analyze a potential upper bound for PTQ quantization since the greedy method is progressive and potentially can correct for errors in the sensitivity ordering. Table~\ref{tab:greedy_aH} shows the resulting latency percentages compared to a 16-bit floating point baseline for ResNet50 and MobileNetV2 given Hessian and augmented Hessian layer ordering. Generally the performance is close to the bisection search (see Table~\ref{tab:resnet50} and~\ref{tab:mbv2}) for both the Hessian and augmented Hessian underlining the contribution of the augmented Hessians even in the greedy setting.  We have to note that the ResNet50 with a 99.9\% target performed worse in a single greedy run than for our bisection search, which we attribute to the progressive nature of the greedy search, e.g. during the search layers are individually added and discarded wherein the bisection search quantizes multiple layers at a time so that cross layer interactions have a chance to compensate for accuracy degradations. Our main takeaway here is that a bisection search with augmented Hessian can perform a quick search (see Table~\ref{tab:eval_num} for number of evaluations) which results in best possible accuracy. Note that the greedy search took at least $N$ evaluations where $N$ is the number of layers which is about $8\times$ more evaluations than the bisection search.

\begin{table}[H]
\caption{Results for a greedy search using Hessians and augmented Hessians on ResNet50 and MobileNetV2. Percentage numbers in the table are relative \textbf{latencies} compared to a 16 bit floating point baseline.}
    \centering
    \begin{tabular}{rcccc}
    \toprule
        Accuracy Target & 97.5\% & 99\% & 99.9\% & 99.99\% \\
    \midrule
        ResNet50 Hessian & 69.49\% & 73.38\% & 81.81\% & 85.73\% \\
        ResNet50  AugHessian & 70.97\% & 71.97\% & 83.21\% & 84.63\% \\ 
        MobileNetV2 Hessian & 75.72\% & 78.37\% & 81.05\% & 83.20\%  \\
        MobileNetV2 AugHessian & 75.72\% &	77.95\%	& 79.91\%	& 86.80\% \\
        \bottomrule
    \end{tabular}
    \label{tab:greedy_aH}
\end{table}

\subsection{Search with Inter-Layer Dependencies}

Given that the inter-layer dependencies augment the Hessian in a meaningful way and improve PTQ we also ran a bisection with only inter-layer dependencies. The resulting latencies for ResNet50, MobileNetV2, and BERT are in Table~\ref{tab:only_id}. Compared to results from Tables~\ref{tab:resnet50},~\ref{tab:mbv2}, and~\ref{tab:bert}  PTQ without any Hessian information results in roughly 10\% slower latencies. Highlighting the importance of combined Hessian and inter-layer information.

\begin{table}[H]
\caption{Relative \textbf{latency} numbers of a bisection search using only inter-layer dependencies for ResNet50, MobileNetV2, and BERT.}
    \centering
    \begin{tabular}{rcccccc}
    \toprule
        Accuracy Target & 95\% & 97.5\% & 98\% & 99\% & 99.9 \%& 99.99\% \\ 
        \midrule
        ResNet50 & 69.92\% & 71.90\% & 83.33\% & 80.67\% & 82.53\% & 83.33\% \\ 
        MobileNetV2 & 75.85\% & 78.62\% & 78.62\% & 78.36\% & 85.68\% & 87.56\% \\ 
        BERT & 73.48\% & 72.27\% & 72.27\% & 72.27\% & 79.16\% & 85.35\% \\
        \bottomrule
    \end{tabular}
    \label{tab:only_id}
\end{table}

\subsection{Sensitivity Threshold}
\begin{figure}[H]
\centering
\includegraphics[width=1.0\textwidth]{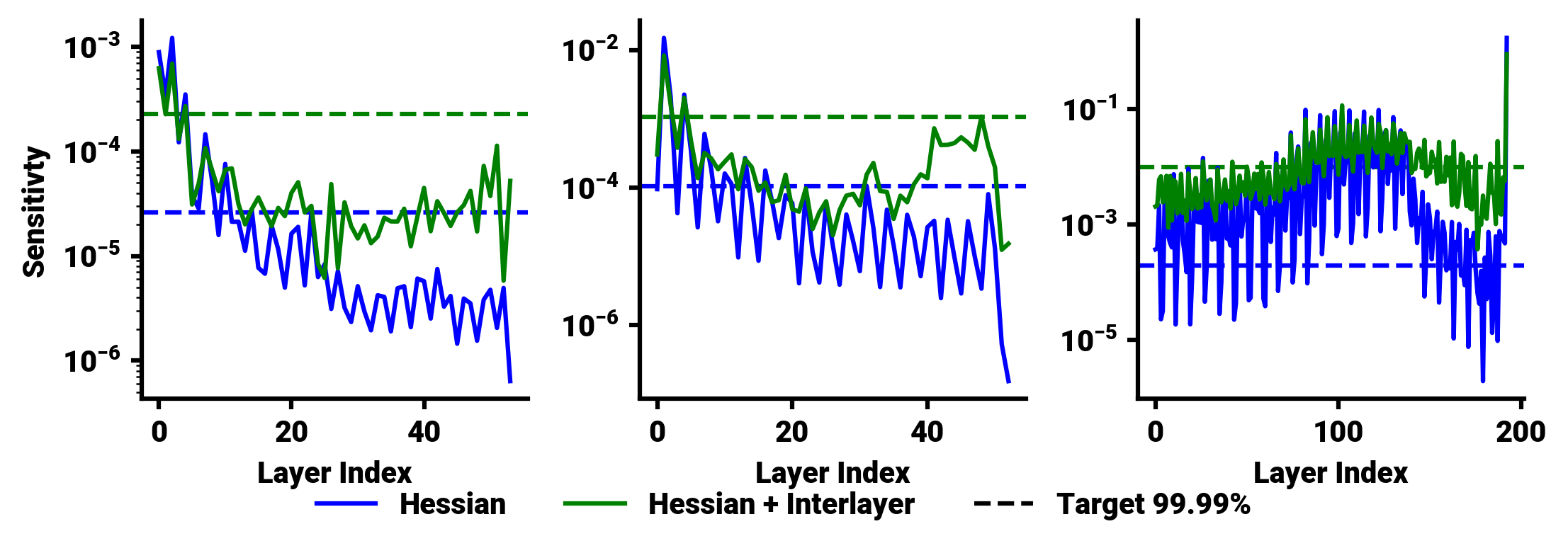}
\caption{ResNet50, MobileNetV2 and BERT layer sensitivity of the Hessian and augmented Hessian. The dashed line shows the final sensitivity threshold for eight bit quantization (e.g. points above the line remain at 16 bits and below are quantized to eight bits or lower) and demonstrating a significant gap between traditional Hessian based metric and our augmented Hessian.}
\label{fig:sens1}
\end{figure}

\subsection{Observed Search Length}

\setlength{\tabcolsep}{2pt}
\begin{table}[H]
\caption{Search iterations required to determine a mixed precision configuration. Our search has two steps: (i) determining all layers quantizable down to eighth bits and (ii) finding all layers which can be quantized down to four bits (from the subset of layers which already can be quantized down to eight). Most search take six configuration evaluations which is in line with the theoretical time complexity of bisection search $\mathcal{O}(\log{N})$, e.g $\log{54} = 5.7549$ (ResNet50), $\log{53}= 5.7280$, and $\log{193} = 7.5925$.}
\label{tab:eval_num}
\begin{center}
\begin{small}
\begin{tabular}{@{}lccccccccccccccccccccccc@{}}
\toprule

& \multicolumn{3}{c}{ResNet50 99\%} & & \multicolumn{3}{c}{ResNet50 99.9\%}	& & \multicolumn{3}{c}{MBV2 99\%} & & \multicolumn{3}{c}{MBV2 99.9\%} & & \multicolumn{3}{c}{BERT 99\%} & & \multicolumn{3}{c}{BERT 99.9\%} \\
\cmidrule{2-4} \cmidrule{6-8}  \cmidrule{10-12} \cmidrule{14-16} \cmidrule{18-20}  \cmidrule{22-24} 
 & 8 bit & & 4 bit & & 8 bit & &  4 bit & & 8 bit & &  4 bit & & 8 bit & &  4 bit & & 8 bit & &  4 bit & & 8 bit & & 4 bit \\
\cmidrule{2-2} \cmidrule{4-4} \cmidrule{6-6} \cmidrule{8-8}  \cmidrule{10-10} \cmidrule{12-12} \cmidrule{14-14} \cmidrule{16-16}  \cmidrule{18-18} \cmidrule{20-20} \cmidrule{22-22} \cmidrule{24-24}
Hessian & 6 & & 5 & & 6 & & 4 & & 6 & & 6 & & 6 & & 6 & & 6 & & 6 & & 6 & & 7 \\
AugHessian & 6 & & 6 & & 6 & & 5 & & 6 & & 6 & & 6 & & 6 & & 6 & & 8 & & 6 & & 7 \\

\bottomrule
\end{tabular}
\end{small}
\end{center}
\end{table}

\subsection{Additional Comparison to Other Work}

\setlength{\tabcolsep}{4pt}
\begin{table}[H]
\caption{Additional comparison to other work for ResNet50. We excluded those models from the main text since all of them focus on four bit quantization and their relative performance degradation is bigger than works mentioned in Table~\ref{tab:resnet50}.}
\label{tab:resnet50_ext}
\begin{center}
\begin{small}
\begin{tabular}{@{}lccccccccccccc@{}}
\toprule
\textbf{ResNet50} & & &  \multicolumn{2}{c}{Accuracy} & & \multicolumn{2}{c}{Size} & & \multicolumn{2}{c}{Latency} & & \multicolumn{2}{c}{Precision} \\
& No FT & & Absolute & Relative$^*$ & &  MB & Relative & & ms & Relative & & W & A\\
\cmidrule{2-2} \cmidrule{4-5} \cmidrule{7-8} \cmidrule{10-11} \cmidrule{13-14} 
Baseline (ours) & & & 77.60 & 100.00\% & & 51.00 & 100.00\% & & 5.20  & 100.00\% & & 16 & 16  \\
\midrule
DFQ \cite{nagel2019data}& \checkmark & & 64.50 & 83.55\%  & & 13.78 & 27.02\%  & & 2.70  & 51.87\%  & & 4 & 4$^\dagger$ \\
ACIQ \cite{banner2018aciq} &  \checkmark & & 68.10 & 88.21\%  & & 13.78 & 27.02\%  & & 2.70  & 51.87\%   & & 4 & 4$^\dagger$ \\
PTQ-MP \cite{liu2021post} & \checkmark &  & 72.67 & 95.43\%  & & 13.78 &  27.02\% &  & 2.70  & 51.87\%   & & 4 & 4$^\dagger$ \\
BRECQ \cite{li2021brecq} & \checkmark &  & 75.05 & 97.22\%  & &13.78 & 27.02\%  & & 2.70  & 51.87\%   & & 4 & 4$^\dagger$ \\
MrBiQ \cite{jeon2022mr} & \checkmark & & 75.17 & 97.62\%  & & 13.78 & 27.02\%  & & 2.70  & 51.87\%   & & 4 & 4$^\dagger$ \\
\bottomrule
\end{tabular}
\end{small}
\end{center}
\end{table}

\subsection{Search Algorithm Pseudo Code}

\begin{algorithm}[H]
  \caption{Bisection search for ideal quantization configuration. Worst and average time complexity is $\mathcal{O}(b\log{N})$ with $b$ as the number of bit-width choices and $N$ the number of layers.}
  \label{alg:bisect}
\begin{algorithmic}[1]
  \State {\bfseries Input:} data $x$, sensitivity metric $s$, accuracy  target $t$, available bit-widths $bs$, model $f$.
  \State Initialize working configuration $w$ with $\max(bs)$.
  \State Initialize layer list $ll$ with all layers of $f$.
  \State Sort $ll$ by $s$ in ascending order.
  \For{$b$ {\bfseries in} $bs$}
      \State Initialize threhsold $thr = \text{length}(ll)/2$.
      \State Initialize upper limit $upl$ to $\text{length}(ll)$.
      \State Initialize lower limit $lowl$ to $0$.
      \Repeat
        \State Initialize local working config $lw$ with $w$.
        \State $lw[ll[0:thr]] \gets b$.
        \State Evaluate $f(x, lw)$ and save accuracy $a$.
        \If{$a >= t$ }
            \State $lowl \gets thr$.
            \State $thr \gets thr+(upl - thr)/2$.
        \Else{}
            \State $upl \gets thr$.
            \State $thr \gets thr - (thr - lowl)/2$.
        \EndIf
      \Until{$thr$ is not changing.}
      \State $w[ll[0:thr]] \gets b$.
      \State $ll \gets ll[0:thr]$.
  \EndFor
  \State {\bfseries Return:} optimal working configuration $w$.
\end{algorithmic}
\end{algorithm}

\begin{algorithm}[H]
  \caption{Progressive approach for ideal quantization configuration. Average time complexity is $\mathcal{O}((2-2^{-(b-1)})N)$ and worst case $\mathcal{O}(bN)$ where $b$ is the number of bit-width choices and $N$ the number of layers. }
  \label{alg:greedy}
\begin{algorithmic}[1]
  \State {\bfseries Input:} data $x$, sensitivity metric $s$, accuracy target $t$, available bit-widths $bs$, model $f$.
  \State Initialize working configuration $w$ with $\max(bs)$.
  \State Initialize layer list $ll$ with all layers of model.
  \State Sort $ll$ by $s$ in ascending order.
  \For{$b$ {\bfseries in} $bs$}
        \State Initialize quantizable layer $ql \gets \emptyset$.
        \For{$l$ {\bfseries in}  $ll$}
            \State $w[l] \gets b$.
            \State Evaluate $f(x, w)$ and save accuracy $a$.
            \If{$a >= t$ }
                \State Append $l$ to $ql$.
            \Else{}
                \State Set $w[l]$ back to last working value.
            \EndIf
        \EndFor
        \State $ll \gets ql$.
  \EndFor 
  \State {\bfseries Return:} optimal working configuration $w$.
\end{algorithmic}
\end{algorithm}

\end{document}